\documentclass[journal]{IEEEtran}
\usepackage{cite}

\ifCLASSINFOpdf
  \usepackage[pdftex]{graphicx}
  \DeclareGraphicsExtensions{.pdf,.jpeg,.png}
\else
   \usepackage[dvips]{graphicx}
   \DeclareGraphicsExtensions{.eps}
\fi
\usepackage{amsmath}
\usepackage{amsfonts}
\usepackage{amsopn}
\usepackage{amsthm}
\usepackage[linesnumbered,ruled]{algorithm2e}
\usepackage{array}
\ifCLASSOPTIONcompsoc
  \usepackage[caption=false,font=normalsize,labelfont=sf,textfont=sf]{subfig}
\else
 \usepackage[caption=false,font=footnotesize]{subfig}
\fi

\newtheorem{defn}{Definition}
\usepackage{enumerate}
\usepackage{stfloats}
\usepackage[table]{xcolor}
\usepackage{url}
\usepackage{microtype}
\usepackage{multirow}

\begin{document}

\title{\huge A Particle Swarm Optimization-based Flexible Convolutional Auto-Encoder for Image Classification}
\author{Yanan~Sun,~\IEEEmembership{Member,~IEEE,}
       ~Bing~Xue,~\IEEEmembership{Member,~IEEE,}
        ~Mengjie~Zhang,~\IEEEmembership{Senior~Member,~IEEE,}
        \\ and Gary~G.~Yen,~\IEEEmembership{Fellow,~IEEE}
\thanks {This work was supported in part by the Marsden Fund of New Zealand Government under Contracts VUW1209, VUW1509 and VUW1615, Huawei Industry Fund E2880/3663, and the University Research Fund at Victoria University of Wellington 209862/3580, and 213150/3662, in part by the National Natural Science Fund of China for Distinguished Young Scholar under Grant 61625204, and in part by the National Natural Science Foundation of China under Grant 61803277.}
\thanks{Yanan Sun, Bing Xue, and Mengjie Zhang are with the School of Engineering and Computer Science, Victoria University of Wellington, PO Box 600, Wellington 6140, New Zealand (e-mails: yanan.sun@ecs.vuw.ac.nz; bing.xue@ecs.vuw.ac.nz; and mengjie.zhang@ecs.vuw.ac.nz).}
\thanks{Gary G. Yen is with the School of Electrical and Computer Engineering, Oklahoma State University, Stillwater, OK 74078 USA (e-mail:gyen@okstate.edu).}
}

\maketitle

\begin{abstract}
Convolutional auto-encoders have shown their remarkable performance in stacking to deep convolutional neural networks for classifying image data during the past several years. However, they are unable to construct the state-of-the-art convolutional neural networks due to their intrinsic architectures. In this regard, we propose a flexible convolutional auto-encoder by eliminating the constraints on the numbers of convolutional layers and pooling layers from the traditional convolutional auto-encoder. We also design an architecture discovery method by exploiting particle swarm optimization, which is capable of automatically searching for the optimal architectures of the proposed flexible convolutional auto-encoder with much less computational resource and without any manual intervention. We test the proposed approach on four extensively used image classification datasets. Experimental results show that our proposed approach in this paper significantly outperforms the peer competitors including the state-of-the-art algorithms.
\end{abstract}

\begin{IEEEkeywords}
Convolutional auto-encoder, particle swarm optimization, image classification, deep learning, neural networks.
\end{IEEEkeywords}

\IEEEpeerreviewmaketitle

\section{Introduction}
\label{section_1}
\IEEEPARstart{A}{uto-Encoders} (AEs)~\cite{rumelhart1988learning,bourlard1988auto,hinton1994autoencoders,schwenk1995transformation} are building blocks of Stacked AE (SAE)~\cite{hinton2006reducing,bengio2007greedy} that is one of the tri-mainstream deep learning algorithms~\cite{lecun2015deep} (i.e., others are Deep Belief Networks (DBN)~\cite{hinton2006fast} and Convolutional Neural Networks (CNNs)~\cite{lecun1998gradient,krizhevsky2012imagenet}). An AE is a three-layer neural network comprising one input layer, one hidden layer, and one output layer, where the number of units in the input layer is identical to that in the output layer. Typically, the transformation from the input layer to the hidden layer is called the encoder, and that from the hidden layer to the output layer refers to the decoder. The encoder extracts the features/representations from the input data, while the decoder reconstructs the input data from the features/representations. By minimizing the divergences between the input data and the reconstruction, one AE is trained. An SAE is stacked by multiple trained AEs for learning hierarchical representations that have gained more remarkable performance than ever before in the field of image classification~\cite{krizhevsky2012imagenet,farabet2013learning,szegedy2015going}.

When image data are fed to the SAE, they must be transformed into the vector-form beforehand, which will change their inherent structures and reduce the consecutive performance in turn. For instance, one image is with the form $I\in \mathbb{R}^{n\times n}$, where the pixel $I_{j,k}$ ($1<j<n, 0<k<n$) has the close distance to the pixel $I_{{j-1},k}$. When $I$ is vectorized to $V\in \mathbb{R}^{n^2}$, the relationship between $I_{j,k}$ and $I_{{j-1},k}$ will be changed and may not be neighbors anymore in $V$. Extensive literatures have shown that adjacent information is a key factor in addressing images related problems~\cite{pluim2003mutual,pass1997comparing,lecun1998gradient,krizhevsky2012imagenet,peng2017bag}. To address this issue, Masci \textit{et al.}~\cite{masci2011stacked} proposed the Convolutional AEs (CAEs), where the image data is directly fed in 2-D form. In CAEs, the encoder is composed of one convolutional layer followed by one pooling layer, and the decoder comprises only one deconvolutional layer. Multiple trained CAEs are stacked to a CNN for learning the hierarchical representations that enhance the final classification performance. Inspired by the advantages of CAEs in addressing data with the original 2-D form, variants of CAEs have been proposed subsequently. For example, Norouzi \textit{et al.}~\cite{norouzi2009stacks} proposed the Convolutional Restricted Boltzmann Machines (RBM)~\cite{hinton2006reducing,smolensky1986information} (CRBM). Lee \textit{et al.}~\cite{lee2009convolutional} proposed the convolutional DBN by stacking a group of trained CRBMs. In addition, Zeiler \textit{et al.}~\cite{zeiler2010deconvolutional,zeiler2011adaptive} proposed the inverse convolutional ones based on the sparse coding schema~\cite{olshausen1997sparse}, which inspired Kavukcuoglu \textit{et al.}~\cite{kavukcuoglu2010learning} to design the convolutional stacked sparse coding for solving object recognition tasks. Recently, Du \textit{et al.}~\cite{du2017stacked} proposed the Convolutional Denoising AE (CDAE) by using Denoising AE (DAE)~\cite{vincent2010stacked} to learn the convolutional filters.

Although experimental results from the CAE and its variants have shown benefits in diverse applications, one major limitation exists in that the architectures of their stacked CNNs are inconsistent with those of state-of-the-art CNNs, such as ResNet~\cite{he2016deep} and VGGNet~\cite{simonyan2014very}. To be specific, because one CAE has one convolutional layer and one pooling layer in the encoder part, the stacked CNN has the same numbers of convolutional layers and pooling layers. However, state-of-the-art CNNs are with non-identical numbers of convolutional layers and pooling layers. Because the architecture of CNN is one key ingredient contributing to the final performance, the restriction on the numbers of convolutional layers and pooling layers of CAEs should be removed. However, choosing the appropriate numbers of convolutional layers and pooling layers is intractable due to the non-differentiable and non-convex characteristics in practice, which is related to the architecture optimization for neural networks.

Algorithms for automatically searching for the optimal architectures of neural networks can be classified into three different categories. The first refers to the algorithms based on the stochastic system, including Random Search (RS)~\cite{bergstra2012random}, Bayesian-based Gaussian Process (BGP)~\cite{rasmussen2006gaussian,movckus1975bayesian}, Tree-structured Parzen Estimators (TPE)~\cite{bergstra2011algorithms}, Sequential Model-Based Global Optimization (SMBO)~\cite{hutter2011sequential}, Evolving Unsupervised Deep Neural Network (EUDNN)~\cite{sun2018evolving},	structure learning~\cite{liu2017structure} and sparse feature learning~\cite{gong2015multiobjective}. The second covers the algorithms that are designed specifically for CNNs where multiple different building blocks exist. The Meta-modeling algorithm (MetaQNN)~\cite{baker2017designing} and the Large Evolution for Image Classification (LEIC) algorithm~\cite{real2017large} belong to this category. The third refers to the NeuroEvolution of Augmenting Topologies (NEAT)~\cite{stanley2009hypercube} algorithm and its diverse variants, such as~\cite{pugh2013evolving,kim2015deep,fernando2016convolution}. Above all, there is one method that does not belong to these categories, i.e., the Grid Search method (GS), which tests every combination of the related parameters.

Particle Swarm Optimization (PSO) is a population-based stochastic evolutionary computation algorithm, motivated by the social behavior of fish schooling or bird flocking~\cite{kennedy1995ieee,eberhart1995new}, commonly used for solving optimization problems without requiring domain knowledge. Compared with other heuristic algorithms, PSO is enriched with the features of the simple concept, easy implementation, and computational efficiency. In PSO, the individuals are called particles, each particle maintains the best solution (denoted by $pBest_i$ for the $i$-th particle) from the memory of itself, and the population records the best solution (denoted by $gBest$) from the history of all particles. During the process, particles expectedly cooperate and interact with the $pBest_i$ and $gBest$, enhancing the search ability and pursuing the optimal solutions. Due to the characteristics of no requirements (e.g., convex or differentiable) imposed on the problems to be optimized, PSO has been widely applied to various real-world applications~\cite{xue2013particle,mohemmed2009particle,setayesh2013novel}, naturally including the architecture design of neural networks, such as~\cite{yu2008evolving,settles2003comparison,da2005improved,juang2004hybrid,lu2003application,salerno1997using}. In the optimization of neural network architectures, these algorithms employ an implicit method to encode each connection of the neural networks and take PSO or its variants to search for the optimum. However, they cannot be utilized for CAEs and CNNs, even SAEs and DBNs, which are deep learning algorithms, where tremendous numbers of connection weights exist, causing the unaffordable cost for implementation and effective optimization in these existing PSO-based architecture optimization algorithms~\cite{omidvar2014cooperative}. As have discussed, CAEs without the constraints on the numbers of the convolutional layer and pooling layers would be greatly preferred for stacking the state-of-the-art CNNs. However, the absolute numbers of these layers are unknown before the architecture is identified. When PSO is employed for the architecture optimization, particles will need to have different lengths. The reasons are that: 1) the length of the particle refers to the number of decision variables of the problem to be solved by PSO; and 2) in the architecture optimization problems, a set of different architectures are involved, and different architectures have different numbers of decision variables. However, the canonical PSO did not provide any way to update the velocity of particles with non-identical lengths. In addition, evaluating particles each of which represents a deep learning algorithm is time-consuming, and will become even more intractable for the population-based updating process. A common way to solve this problem is to employ intensive computational resources and utilize parallel-computation techniques.

The objective of this paper is to design and develop an effective and efficient PSO method to automatically discover the architecture of the flexible convolutional auto-encoder without manual intervention. To achieve this goal, we have specified the four aims as follows:

\begin{enumerate}
	\item  Propose a Flexible CAE (FCAE) where multiple convolutional layers and pooling layers can exist. The FCAE has no requirement on the particular numbers of the convolutional layers and the pooling layers, and have the potential for stacking to different types of CNNs.
		
	\item Design a PSO-based Architecture Optimization (PSOAO) algorithm for the proposed FCAE. In PSOAO, we will propose an efficient encoding strategy to represent the FCAE architectures, which involve hundreds of thousands of parameters, into each particle, and we will also develop an effective velocity updating mechanism for particles with variable lengths.
	
	\item Investigate the performance of the proposed FCAE when its architecture is optimized by the designed PSOAO on image classification benchmark datasets (i.e, the CIFAR-10 dataset~\cite{krizhevsky2009learning}, the MNIST dataset~\cite{lecun1998gradient}, the STL-10 dataset~\cite{coates2011analysis}, and the Caltech-101 dataset~\cite{fei2007learning}.), compare the classification accuracy to peer competitors and examine the evolution effectiveness of PSOAO.
	
	\item Investigate the effectiveness of the designed velocity updating method through quantitative experiments on the comparisons to its opponents.
\end{enumerate}

 The remainder of this paper is organized as follows. Background of the CAE and PSO is reviewed in Section~\ref{section_2}. This is followed by the details of the proposed PSOAO algorithm in Section~\ref{section_3}. Then, the experiment design and the result analysis are documented in Sections~\ref{section_4} and~\ref{section_5}, respectively. Finally, the conclusions and future work are drawn in Section~\ref{section_7}.
\section{Literature Review}
This work will build FCAE and PSOAO based on CAE and PSO, respectively. Therefore, we would like to provide the skeletons of CAE and PSO as well as their limitations for FCAE in the following subsections, which could help the readers to conveniently understand our work in this paper. In addition, related works are also reviewed and commented, which helps the readers to easily appreciate the importance of our work in this paper.
\label{section_2}
\subsection{Convolutional Auto-Encoder}
\label{section_2_1}

\begin{figure}
	\centering
	\includegraphics[width=0.8\columnwidth]{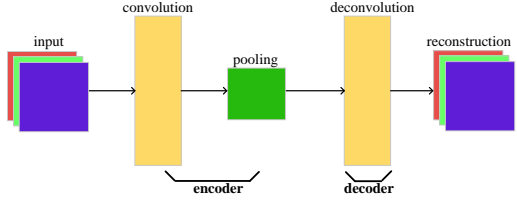}
	\caption{The illustrative architecture of CAE.}
	\label{fig_cae_architecture}
\end{figure}
For the convenience of the development, assuming CAEs are utilized for image classification tasks, and each image $X\in \mathbb{R}^{w\times h \times c}$, where $w$, $h$, $c$ refer to the image width, height, and number of channels, respectively. Fig.~\ref{fig_cae_architecture} illustrates the architecture of one CAE~\cite{masci2011stacked}.

\textbf{\textit{Convolution}:} Given the input data, the convolution operation employs one filter to slide with one defined stride, and outputs the element that is the sum of the products of the filter and the input data with which this filter overlaps. All elements generated by one filter construct one feature map, and multiple feature maps are allowed in the convolution operations. The convolution operation has the SAME type and the VALID type. The parameters related to the convolutional operation are the \textit{filter size} (width and height), the \textit{stride size} (width and height), the \textit{convolutional type}, and the \textit{number of feature maps}.

\textbf{\textit{Pooling}:} Pooling operation resembles the convolution operation, in addition to the filter and the way to generate the elements of the corresponding feature map. Specifically, the filter in a pooling operation is called a ``kernel'', and no value exists in the kernel. There are two types of statistical indicators in the pooling operation: mean and maximal. Typically, maximal pooling is preferred in CAEs.The pooling operation requires the parameters: the \textit{kernel size} (width and height), the \textit{stride size} (width and height), and the \textit{pooling type}.

\textbf{\textit{Deconvolution}:} The deconvolutional operation is equivalent to the corresponding convolutional operation with inverse parameter settings. Specifically, the deconvolutional operation performs the convolutional operation with the filter and stride, which has been used in the corresponding convolutional operation, on the feature map that is resulted from the corresponding convolutional operation. In order to assure the output of the deconvolutional operation to have the same size as the input of the corresponding convolutional operation, extra zeros may be padded to the input of the deconvolutional operation.

\textbf{\textit{Learning of CAE}:} The mathematical form of the CAE is represented by Equation~(\ref{equ_cae}), where the $conv(\cdot)$, $pool(\cdot)$, and $de\_conv(\cdot)$ denote the convolution, pooling, and deconvolution operations, respectively, $F(\cdot)$ and $G(\cdot)$ refer to the element-wise nonlinear activation functions, $b_1$ and $b_2$ are the corresponding bias terms, $r$ and $\hat{X}$ are the learned features and reconstruction of $X$, $l(\cdot)$ measures the differences between $X$ and $\hat{X}$, and $\Omega$ is the regularization term to improve the feature quality. By minimizing $L$, the CAE is trained, and then parameters in convolution operation, bias terms, and deconvolutional operation are identified. Encoders with these parameters from multiple trained CAEs are composed to be a CNN for learning hierarchical features that benefit the final classification performance~\cite{lecun2015deep}.

\begin{equation}
\label{equ_cae}
\left\{
\begin{array}{l}
\begin{array}{ll}
r =& pool\left (F\left (conv\left (X\right )+b_1\right )\right ) \\[5pt]
\hat{X}=&G\left(de\_conv\left(r\right) + b_2\right) \\[5pt]
\end{array}
\\
 minimize~~~L=l(X, \hat{X}) + \Omega
\end{array}
\right.
\end{equation}

\subsection{Motivation of FCAE}
As shown in Fig.~\ref{fig_cae_architecture}, a CAE is composed of a convolutional layer followed by a pooling layer and then a deconvolutional layer. The transformation through the convolutional layer and then the pooling layer is called the encoder. The transformation through the deconvolutional layer is called the decoder. During the construction of a CNN by using CAEs, encoders from multiple CAEs are stacked together based on their training orders, and the input data of the current encoder is the output data of the previous one. However, this architecture is not able to form the state-of-the-art CNNs nor the common deep CNNs. Next, we will describe them in detail.

\begin{figure}[htp]
	\centering
	\subfloat[]{\includegraphics[width=0.9\columnwidth]{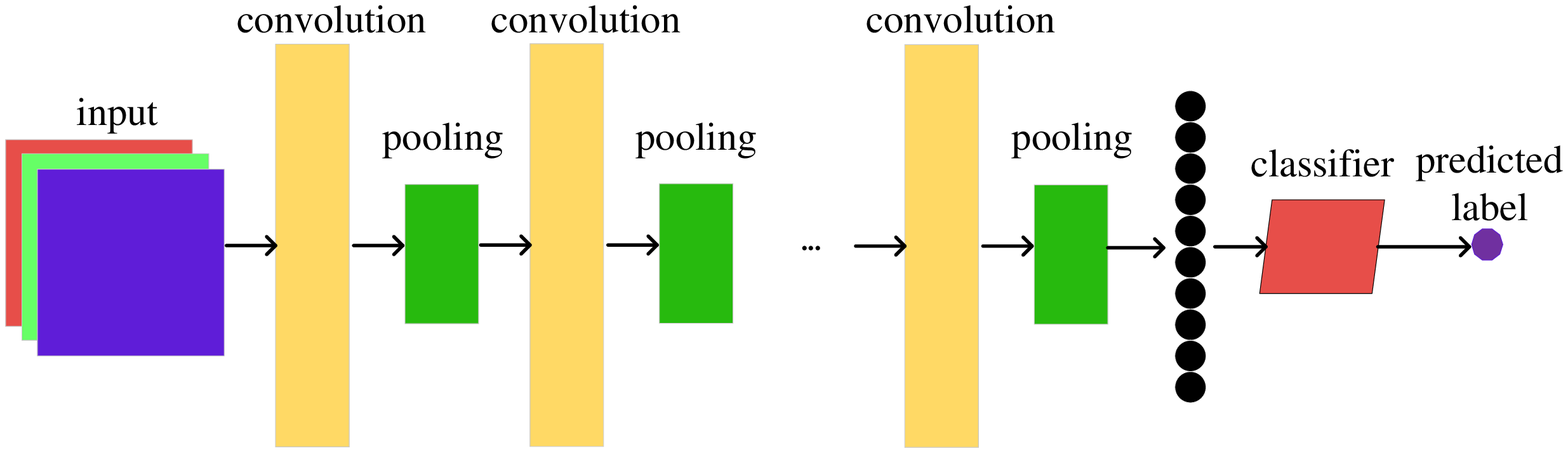}%
		\label{fig_cnn_architecture}}
	\hfill
	\subfloat[]{\includegraphics[width=0.9\columnwidth]{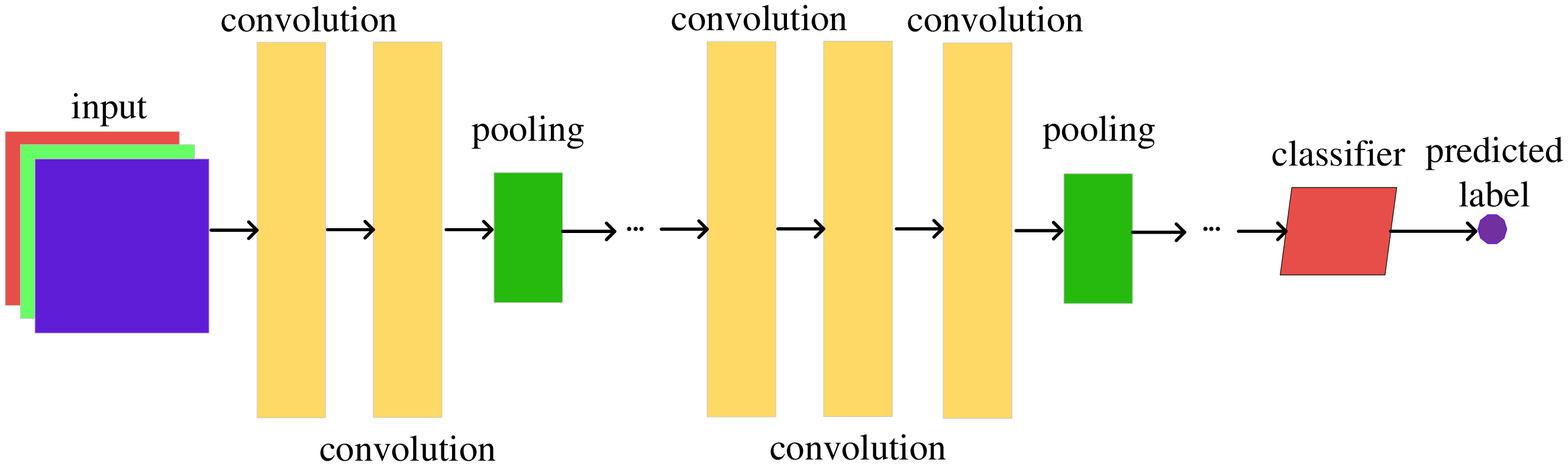}%
		\label{fig_vgg_architecture}}

	\caption{Architectures of the CNN stacked by CAEs (in Fig.~\ref{fig_cnn_architecture}) and VGGNet (in Fig.~\ref{fig_vgg_architecture}).}
	\label{fig_architecture_comparison}
\end{figure}

The architectures of a CNN stacked by CAEs and a state-of-the-art CNN named VGGNet~\cite{simonyan2014very} are shown in Fig.~\ref{fig_cnn_architecture} and Fig.~\ref{fig_vgg_architecture}, respectively. From these two examples, it is evident that CAEs are incapable of stacking into VGGNet. The reason is that the CNN stacked by CAEs are with the same building blocks, i.e., the two-layer network including a convolutional layer followed by a pooling layer. Consequently, the stacked CNN is composed of a series of such building blocks and with the same number of the convolutional layers to that of the pooling layers. In addition, a convolutional layer must be followed by a pooling layer in the CNNs stacked by CAEs. As can be observed from Fig.~\ref{fig_vgg_architecture}, there are multiple convolutional layers and pooling layers with the identical numbers, and also a convolutional layer is not necessarily followed by a pooling layer.

The pooling layer is typically used to reduce the dimension of the input data to decrease the computational complexity. The most commonly used configuration for a pooling layer is with the kernel size of $2\times 2$ and the stride of $2\times 2$. Based on the working mechanism introduced in Subsection~\ref{section_2_1}, one pooling layer with such a configuration will reduce half size of the input data. For example, for an image from the CIFAR-10 dataset~\cite{krizhevsky2009learning} that is with the dimension of $32\times 32$, the output size will become $1\times 1$ with 5 pooling layers, i.e., the CNN designed to process CIFAR-10 with CAEs is at most up to $10$ layers. However, the state-of-the-art designs on this benchmark have more than $100$ layers~\cite{huang2017densely}.

Based on the description above, the root of the limitations from CAE exists in its architectures that the encoder of a CAE is composed one convolutional layer and then one pooling layer. Therefore, the concern is naturally raised that the architecture of CAE should be revised, which motivates the design of FCAE, i.e., in the encoder part of an FCAE, non-identical numbers of convolutional layers and pooling layers are allowed, and a pooling layer can follow a series of convolutional layers.

\subsection{Particle Swarm Optimization}
\label{section_2_2}
A typical PSO has the procedure as follows:
\begin{enumerate}[Step 1):]
	\item Initialize the particles, predefine a maximal generation number $max_t$, and initialize a counter $t=0$;
	\item Evaluate the fitness of particles;\label{pso_simple_step2}
	\item For each particle, choose the best one, $pBest_i$, from its memory;
	\item Choose the best particle $gBest$ from the history of all particles;
	\item Calculate the velocity $\{v_1,\cdots,v_i,\cdots\}$ of each particle $\{x_1,\cdots,x_i,\cdots\}$ by Equation (\ref{equ_veclocity_calculate});
	\item Update the position $\{p_1,\cdots,p_i,\cdots\}$ of each particle $\{x_1,\cdots,x_i,\cdots\}$ by Equation (\ref{equ_position_update});\label{pso_simple_step5}
	\item Increase $t$ by $1$, if $t< max_t$ to repeat Steps~\ref{pso_simple_step2})~--~\ref{pso_simple_step5}) otherwise go to Step 8);
	\item The position of $gBest$ is reported.
\end{enumerate}

\begin{equation}
\label{equ_veclocity_calculate}
v_i \leftarrow \overbrace{w\cdot v_i}^{\text{inertia}} + \overbrace{c_1\cdot r_1 \cdot (p_g - p_i)}^{\text{global search}}   + \overbrace{c_2\cdot r_2\cdot (p_p - p_i)}^{\text{local search}}
\end{equation}

\begin{equation}
\label{equ_position_update}
p_i\leftarrow p_i + v_i
\end{equation}

In Equation~(\ref{equ_veclocity_calculate}), $w$ denotes the inertia weight, $c_1$ and $c_2$ are acceleration constants, $r_1$ and $r_2$ are random numbers between $0$ and $1$, and $p_g$ as well as $p_p$ denotes the positions of $gBest$ as well as $pBest_i$, respectively. $v_i$ and $p_i$ denote the velocity and position of the $i$-th particle $x_i$, respectively. By integrating the ``inertia'', ``global search'', and ``local search'' terms into the velocity updating, the best position is expected to be found by particles.

Noting in Equation~(\ref{equ_veclocity_calculate}) that there are two subtraction operations existing in the ``global search'' and ``local search''. In order to better understand how such a subtraction operation works, we will describe the details by taking the term $p_g-p_i$ in ``global search'' as an example. Supposing the minimization problem to be solved is formulated as $f(z_1,z_2,\cdots,z_n)$ where there are $n$ decision variables $\{z_1,z_2,\cdots,z_n\}\in \Psi$. When PSO is used to solve this problem, the $i$-th particle $x_i$ will be randomly sampled from $\Psi$, and its position is determined by a particular value $p_i=\{z_1^i, z_2^i, \dots, z_n^i\}\in \Psi$. Through the interaction formulated by Equations~(\ref{equ_veclocity_calculate}) and~(\ref{equ_position_update}), the position of $x_i$ is updated towards the position of the optimal solution. After a number of iterations, the optimization is solved and the final solution is the position of the global best particle, $gBest$. In this example, the length of a particle is $n$, i.e., the number of decision variables and also the dimension of the position. Obviously, $p_g-p_i=\{z_1^g-z_1^i,z_2^g-z_2^i,\cdots,z_n^g-z_n^i\}$.

\textbf{\textit{Limitation of Using PSO for Architecture Design in FCAE}:} The velocity updating requires the particle $x_i$, $gBest$, and $pBest_i$ to have the same/fixed length. When PSO is used for the architecture optimization of FCAE, particles represent the potential optimal architectures of FCAE. Because the optimal architecture of FCAE for solving the task at hand is unknown, particles with different variable lengths will emerge.  Therefore, a novel velocity updating method needs to be designed in this regard.

\subsection{Related Works}
\label{section_2_4}
As we have discussed in Section~\ref{section_1}, algorithms for optimizing the architectures of deep neural networks fall into four different categories.

For the algorithms in the first category, RS~\cite{bergstra2012random} evaluates the randomly selected architectures with a predefined maximal trial number, and uses the best one in performance. RS has been reported that it only works under the condition when optimum is in a subspace of the whole search space~\cite{atanassov2009tuning}. However, it is not clear whether this fact equally applies to CNNs or not. Compared to RS, BGP~\cite{rasmussen2006gaussian,movckus1975bayesian} utilizes more knowledge on choosing the potential optimal architecture based on Bayesian inference~\cite{movckus1975bayesian}. However, BGP has extra parameters, such as the kernels, which are hard to tune. TPE~\cite{bergstra2011algorithms} works with the assumption that the parameters related to the architecture are independent, while most parameters in CNNs are indeed dependent such as the convolutional layer size and the strides. In addition, EUDNN is designed for unsupervised deep neural networks that are with different architectures from CNNs. Furthermore, methods in~\cite{liu2017structure,gong2015multiobjective} are only used for optimizing the particular network connections with the given architectures, such as the sparsity. To this end, algorithms in this category cannot be used to optimize the architectures of CNNs. Consequently, they are not suitable for FCAEs of which the architectures are based on CNNs.

For the algorithms in the second category, i.e., MetaQNN~\cite{baker2017designing} and LEIC~\cite{real2017large}, both of them are designed specifically for optimizing the architectures of CNNs. Specifically, MetaQNN uses the reinforcement learning technique~\cite{sutton1998reinforcement} to heuristically exploit the potential optimal architectures of CNNs, thoroughly evaluate them, and then choose the one that has the best fitness. LEIC employs nearly the same strategy to MetaQNN except for that LEIC used genetic algorithm as the heuristic method. Due to the complete training on each candidate in both algorithms, their deficiencies are also obvious, i.e., their training relies on the extensive computational resources. For instance, MetaQNN employed 10 Graphics Processing Unit (GPU) cards for 8-10 days, while LEIC employed 250 high-performance computers for 20 days on the CIFAR-10 test problem~\cite{krizhevsky2009learning}. Unfortunately, sufficient computational resource is not necessarily available to all interested researchers.

Because the algorithms in the third category are all based on NEAT, their working flows are nearly the same. Specifically, the input layer and the output layer are assigned first, and then neurons between these two layers and connections from arbitrary two neurons are heuristically generated. With the fitness evaluation upon each situation, the better ones are selected and the best one is expected to be found with genetic algorithm. In addition to the limitations occurred in the algorithms from the second category, other limitations from NEAT-based algorithms are that hybrid connections (i.e., the weight connections between the layers which are not adjacent) would be produced, and the configurations of the input layer and the output layer must be specified in advance, which are not allowed or applicable in CAEs.

Theoretically, GS can find the optimal architecture because of its exhaustive nature to try each candidate. However, it is impossible for GS to try each candidate in practice. Recently, experimental investigation~\cite{suncomp} shows that GS is only suitable for the problems with no more than four parameters in practice. As have been shown in Subsection~\ref{section_2_1}, a CAE will have more than 10 parameters even it contains only one convolutional layer and one pooling layer. In addition, GS cannot well handle parameters with continuous values because of the ``interval'' problems~\cite{suncomp}.

\section{The proposed PSOAO Algorithm for FCAE}
\label{section_3}
In this section, the details of the proposed PSOAO algorithm for FCAE will be provided. We will describe the encoding strategy which involves the FCAE representation, the particle initialization, the fitness evaluation, and the velocity and position updating mechanism.
\subsection{Algorithm Overview}
\label{sec_3_1}
\begin{algorithm}
	\label{alg_framework}
	\caption{Framework of the PSOAO Algorithm}
	$\textbf{x} \leftarrow$Initialize the particles based on the proposed encoding strategy;\\
	\label{alg_line_1}
	$t\leftarrow 0$;\\
	\While{$t<$ the maximal generation number}
	{\label{alg_begin_evo}
		Evaluate the fitness of each particle in $\textbf{x}$;\\
		\label{alg_line_4}
		Update the $pBest_i$ and $gBest$;\\
		\label{alg_line_5}
		Calculate the velocity of each particle;\\
		\label{alg_line_6}
		Update the position of each particle;\\
		\label{alg_line_7}
		$t\leftarrow t+1$;
	}\label{alg_end_evo}
	\textbf{Return} $gBest$ for deep training.
	\label{alg_line_10}
\end{algorithm}

	Algorithm~\ref{alg_framework} outlines the framework of the designed PSOAO algorithm. Firstly, particles are randomly initialized based on the proposed encoding strategy (line~\ref{alg_line_1}). Then, particles start to evolve until the generation number exceeds the predefined one (lines~\ref{alg_begin_evo}-\ref{alg_end_evo}). Finally, the $gBest$ particle is picked up for obtaining the final performance through the deep training\footnote{Supposing the optimal performance of an neural network-based model is achieved through $T^*$ training epochs. The deep training refers to the model has experienced $T$ training epochs, where $T$ is equal to or greater than $T^*$.} to solve tasks at hand (line~\ref{alg_line_10}).
	
	During the evolution, the fitness of each particle is evaluated (line~\ref{alg_line_4}) first, and then the $pBest_i$ and $gBest$ are updated based on the fitness (line~\ref{alg_line_5}). Next, the velocity of each particle is calculated (line~\ref{alg_line_6}) and their positions are updated (line~\ref{alg_line_7}) for the next generation of evolution. In the following subsections, keys aspects of PSOAO are detailed.

\subsection{Encoding Strategy}
For the convenience of the development, the definition of FCAE is given in Definition~\ref{def_fcae} by generalizing the building blocks in all CNNs. Obviously, the CAE is a special form of FCAE when the numbers of convolutional layers and pooling layers are both set to be $1$.
\begin{defn}
	\label{def_fcae}
	{A flexible convolutional auto-encoder (FCAE) encompasses one encoder and one decoder. The encoder is composed of the convolutional layers and pooling layers, where these two types of layers are not mixed and their numbers are flexible. The decoder part is the inverse form of the encoder.}
\end{defn}

\begin{figure}[t]
	\centering
	
	\includegraphics[width=0.9\columnwidth]{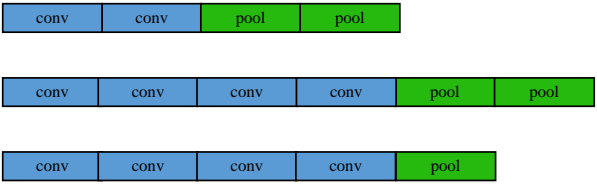}
	\caption{Three particles with different encoded information from PSOAO.}
	\label{fig_particles_architecture}

\end{figure}

We design an encoding strategy through \textit{variable-length} particles to encode the potential architecture of one FCAE into one particle. Because the decoder part in an FCAE is the inverse form of the encoder, the particle in the proposed \textit{variable-length} encoding strategy only encodes the encoder part for the reason of reducing the computational complexity. Each particle contains different numbers of convolutional layers and pooling layers. Based on the introduction of the convolution operation and pooling operation in Subsection~\ref{section_2_1}, all the encoded information of PSOAO for FCAE are summarized in Table~\ref{table_encoded_information}  where the $l_2$ denotes the weight decay regularization term for preventing from the overfitting problem~\cite{krogh1992simple}. Because only the convolutional layers involve weight parameters, this regularization term is applied only to the convolutional layers. Furthermore, because the output size will not change from the input size with the SAME convolutional layers, which is easy to control in automatic architecture discovering, the designed encoding strategy will not encode the type of the convolutional layers but default to the SAME type. As have mentioned in Subsection~\ref{section_2_1} that CAE prefers to the max pooling layer, we also don't need to encode this parameter. In addition, three examples of the generally encoded particles in PSOAO are illustrated in Fig.~\ref{fig_particles_architecture}. In the following, we will detail the rationale of this encoding strategy.

\begin{table}[t]
	\renewcommand{\arraystretch}{1.7}
	\caption{Encoded information in the convolutional layers and the pooling layers of FCAE.}
	\label{table_encoded_information}
	\begin{center}
		\begin{tabular}{p{0.2\columnwidth}<{\centering}p{0.7\columnwidth}<{\centering}}
			\hline
			
			\hline
			\textbf{Layer Type} & \textbf{Encoded Information }\\
			\hline
			
			\hline
			convolutional layer & filter width, filter height, stride width, stride height, convolutional type, number of feature maps, and the coefficient of $l_2$. \\
			\hline
			pooling layer & kernel width, kernel height, stride width, stride height, pooling type\\
			\hline
			
			\hline
		\end{tabular}
	\end{center}
\end{table}

In the proposed PSOAO algorithm, a variable-length encoding strategy is designed for the particles representing FCAEs with different architectures. The major reason is that the optimal architecture is unknown prior to the optimization, and the fixed-length encoding strategy often imposing constraints on architectures does not work under this occasion. Specifically, if the traditional fixed-length encoding strategy is employed, the maximal length should be specified in advance. However, the maximal length is not easy to set and needs to be carefully tuned for the best performance. A too small number denoting the maximal length would be inefficient for the optimized architecture of FCAE to solve complex problems. A too large number would consume much more unnecessary computation, and also results in worse performance within the same predefined evolution generation number. Furthermore, two types of layers exist in each particle, which increases the difficulty of employing the fixed-length encoding strategy. With the designed variable-length encoding strategy, all the information of potential optimal architecture for FCAE can be flexibly represented for exploitation and exploration during the search process without manual intervention.

\subsection{Particle Initialization}
\begin{algorithm}
	\label{alg_particle_init}
	\caption{Particle Initialization}
	\KwIn{The population size $N$, the maximal number of convolutional layers $N_{c}$, and the maximal number of pooling layers $N_{p}$.}
	\KwOut{Initialized population $\textbf{x}_0$.}
	$\textbf{x}\leftarrow\emptyset$;\\
	\While{$|\textbf{x}| \leq N$}
	{
		$conv\_list\leftarrow \emptyset$;\\
		\label{alg_begin_conv_part}
		$n_{c}\leftarrow$ Uniformaly generate an integer between $[1, N_{c}]$;\\
		\While{$conv\_list \leq n_{c}$}
		{
			$conv\_unit\leftarrow$ Initialize a convolutional layer with random settings;\\
			$conv\_list\leftarrow$ $conv\_list\cup conv\_unit$;
		}\label{alg_end_conv_part}
		
		$pool\_list\leftarrow \emptyset$;\\
		\label{alg_begin_pool_part}
		$n_{p}\leftarrow$ Uniformaly generate an integer between $[1, N_{p}]$;\\
		\While{$|pool\_list| \leq n_{p}$}
		{
			$pool\_unit\leftarrow$ Initialize a pooling layer with random settings;\\
			$pool\_list\leftarrow pool\_list\cup pool\_unit$;
		}
		\label{alg_end_pool_part}
		Use $conv\_list$ and $ pool\_list$ to generate a particle $x$;\\
		$\textbf{x}\leftarrow \textbf{x}\cup x$;
	}
	\textbf{Return} $\textbf{x}$.
\end{algorithm}

Algorithm~\ref{alg_particle_init} shows the procedure of the particle initialization with the given population size, and maximal numbers of the convolutional layers and the pooling layers. Particularly, lines~\ref{alg_begin_conv_part}-\ref{alg_end_conv_part} demonstrate the initialization of the convolutional layers, while lines~\ref{alg_begin_pool_part}-\ref{alg_end_pool_part} show the initialization of the pooling layers, where the random settings refer to the settings of the information encoded in these two types of layers. Because the decoder part of FCAE can be explicitly derived from its encoder part, each particle in the proposed PSOAO algorithm contains only the encoder part for reducing the computational complexity.

\subsection{Fitness Evaluation}
\label{sec_fitness}
Algorithm~\ref{alg_particle_evaluation} shows the fitness evaluation for the particles in PSOAO. As we have introduced in Subsection~\ref{section_2_1}, the reconstruction error added by the loss of the regularization term is identified as the objective function for training CAE. However, the loss of the regularization term used in FCAE (i.e., the $l_2$ loss) is highly affected by the weight numbers and weight values, and different architectures have different weight numbers and the weight values. In order to investigate when only the architecture is reflected by the particle quality, the $l_2$ loss is discarded and only the reconstruction error is employed as the fitness. Supposing the batch training data is $\{d_1,d_2,\cdots,d_n\}$ ($d_i^{jk}\in R^{w\times h}$ denotes the pixel value at the position of $(j,k)$ of the $i$-th image in the batch training data, and each image has the dimension of $w\times h$), the weights in the FCAE is $\{w_1,w_2,\cdots,w_m\}$, the reconstructed data is $\{\hat{d_1}, \hat{d_2}, \cdots, \hat{d_n}\}$. The $l_2$ is calculated by $\sum_{i=1}^m w_i^2$, while the reconstruction error is calculated by $\frac{1}{n}\sum_{k=1}^n\sum_{l=1}^w\sum_{m=1}^h(\hat{d_k^{lm}} - d_k^{lm})^2$.

\begin{algorithm}
	\label{alg_particle_evaluation}
	\caption{Fitness Evaluation}
	\KwIn{The population $\textbf{x}$, the training set $D_{train}$, the number $N_{train}$ of training epoch.}
	\KwOut{The population $\textbf{x}$ with fitness.}
	Calculate the reconstruction error and $l_2$ loss of the FCAE encoded by each particle in $\textbf{x}$, and train the weights with $N_{train}$ epochs;\\
	Calculate the reconstruction error of each batch data in $D_{train}$ and set the mean reconstruction error as the fitness of the corresponding particle;\\
	\textbf{Return} $\textbf{x}$.
\end{algorithm}

Typically, a deep learning algorithm requires a training epoch number in the magnitude of $10^2-10^3$ to train its weight parameters by gradient-based algorithms. This high computational issue is even worsen in population-based algorithms. In the proposed PSOAO algorithm, this number is specified at a very smaller number (e.g., $5$ or $10$) for speeding up the training. For example, it will take $2$ minutes for training one epoch on the CIFAR-10 dataset (with $50,000$ training samples) utilizing one GPU card with the model number of GTX1080. If we train it with $10^2$ epochs for each particle with the population size of $50$ for $50$ generations, it will take about one year, which is not acceptable for the purpose of general academic research. The widely used solution for easing this adversity is to employ intensive computation resources, such as the LEIC algorithm very recently proposed by Google in 2017, where $250$ computers are employed for about $20$ days on the CIFAR-10 dataset using the genetic algorithm for the architecture discovering. In fact, it is not necessary to evaluate the final performance of each particle by a large number of training epochs during the architecture searching. Instead picking up a promising particle after a fewer training epochs and then deep training it once with sufficient training epochs could be a promising alternative. In the proposed PSOAO algorithm, a small number of training epochs is employed to conduct the fitness evaluation of particles. With the evaluated fitness, the $gBest$ and $pBest_i$ are selected to guide the search towards the optimum. When the evolution is terminated, the $gBest$ is selected and one-time deep training is performed for reaching the optimal performance. We have shown that this setting can largely speed up PSOAO yet with less computational resources, while the promising performance of PSOAO is still maintained. Specifically, the running time on the investigated benchmark datasets are shown in Table~\ref{table_comsuming_time_alg}, the adopted computational resources for the investigated benchmark datasets are given in Subsection~\ref{sec_4_3}, and the experimental results regarding the performance are shown in Subsections~\ref{sec_5_1} and \ref{sec_5_3}. 

\subsection{Velocity Calculation and Position Update}
In PSOAO, the particles are with different lengths, and Equation~(\ref{equ_veclocity_calculate}) cannot be directly used. To solve this problem, we design a method named ``$x$-reference'' to update the velocity. In $x$-reference, the lengths of $gBest$ and $pBest_i$ refer to the length of the current particle $x$, i.e., $gBest$ and $pBest_i$ keep the same length to that of $x$. Because the $pBest_i$ is selected from the memory of each particle, and the $gBest$ is chosen from all particles, the current particle $x$ is always with the same length of $pBest_i$ and the $x$-reference is applied only to the ``global search'' part of Equation~(\ref{equ_veclocity_calculate}). Algorithm~\ref{alg_particle_velocity} shows the details of the $x$-reference method.

\begin{algorithm}[htp]
	\label{alg_particle_velocity}
	\caption{The $x$-reference Velocity Calculation Method}
	\KwIn{The partilce $x$, the $gBest$, the acceleration constant $c_1$.}
	\KwOut{The gloabl search part of velocity updating.}
	$r_1\leftarrow$ Randomly sample a number from $[0,1]$;\\
	$cg\leftarrow$ Extract the convolutional layers from $gBest$;\\
	\label{alg_vc_begin}
	$cx\leftarrow$ Extract the convolutional layers from $x$;\\
	$pos\_c\leftarrow\emptyset$;\\
	\uIf {$|cg|<|cx|$}
	{
		$c\leftarrow$ Initialize $|cx|-|cg|$ convolutional layers with encoded information of zeros;\\
		$cg\leftarrow$ Pad $c$ to the tail of $cg$;\\
	}\Else{
		$cg\leftarrow$ Truncate the last $|cg|-|cx|$ convolutional layers from $cg$;
	}
	\For{$i = 1~\text{to}~|cx|$}
	{
		$p_{cg_i}, p_{x_i}\leftarrow$ Extract the position of the $i$-th convolution layer from $cg$ and $cx$;\\
		$pos\_c\leftarrow pos\_c\cup c_1\cdot r_1\cdot (p_{cg_i} - p_{x_i})$;\\
	}\label{alg_vc_end}
	
	$pos\_p\leftarrow$ Analogy the operations in lines~\ref{alg_vc_begin}-\ref{alg_vc_end} on the pooling layers of $gBest$ and $x$;\\
	\label{alg_vc_pooling}
	
	\textbf{Return} $pos\_c\cup pos\_p$.
\end{algorithm}

Specifically, the $x$-reference method is applied twice in the same manner in the velocity updating for the ``global search'' part of Equation~(\ref{equ_veclocity_calculate}). The first is on the convolutional layer part of $gBest$ and $x$ (lines~\ref{alg_vc_begin}-\ref{alg_vc_end}), and the second is on the pooling layer part (line~\ref{alg_vc_pooling}). For the convolutional layer part, if the number of convolutional layers, $cg$, from $gBest$ is smaller than that from $x$, new convolutional layers initialized with zero values are padded to the tail of $cg$. Otherwise, extra convolutional layers are truncated from the tail of $cg$. After the ``global search'' part is derived by Algorithm~\ref{alg_particle_velocity}, the ``inertia'' and ``local search'' parts in Equation~(\ref{equ_veclocity_calculate}) are calculated as normal, then the complete velocity is calculated and the particle position is updated by Equation~(\ref{equ_position_update}).

\begin{figure*}
	\centering
	\subfloat[]{\includegraphics[width=1.8\columnwidth]{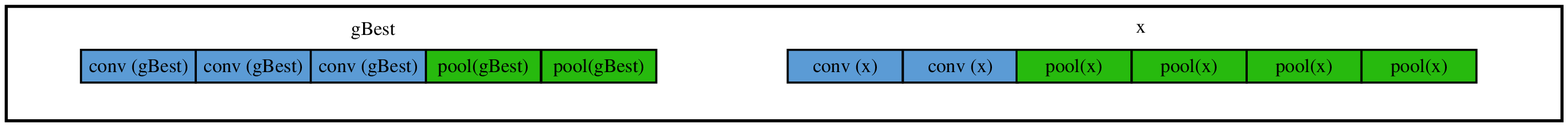}%
		\label{fig_velocity_1}}
	\hfill
	\subfloat[]{\includegraphics[width=1.8\columnwidth]{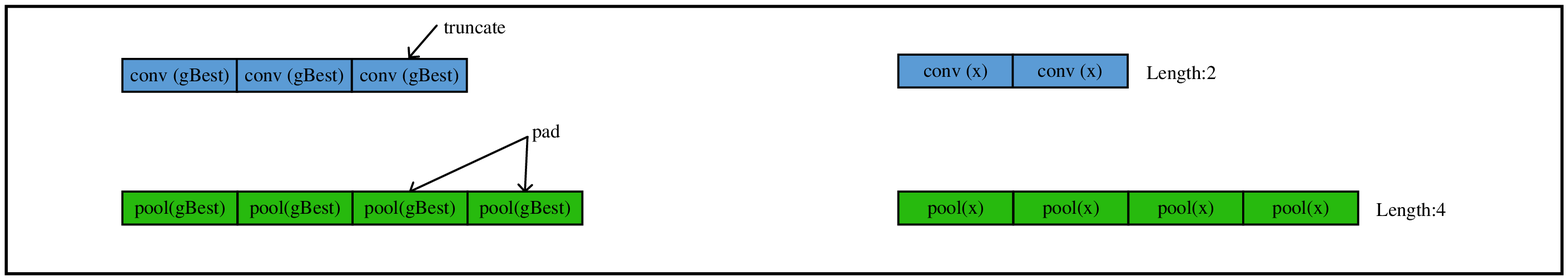}%
		\label{fig_velocity_2}}
	\hfill
	\subfloat[]{\includegraphics[width=1.8\columnwidth]{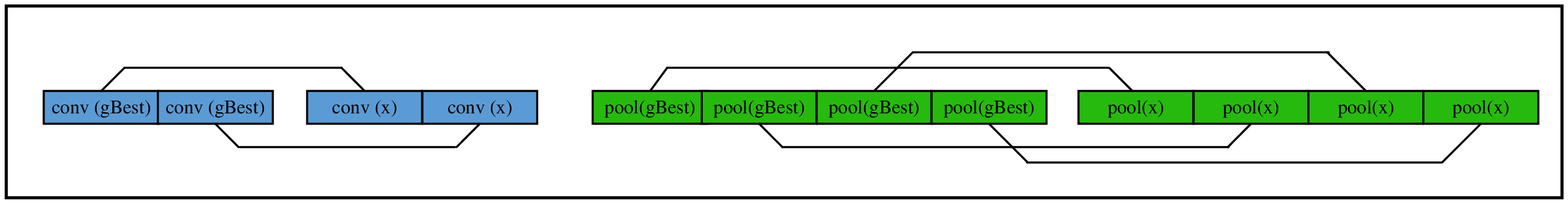}%
		\label{fig_velocity_3}}
	\hfill
	\subfloat[]{\includegraphics[width=1.8\columnwidth]{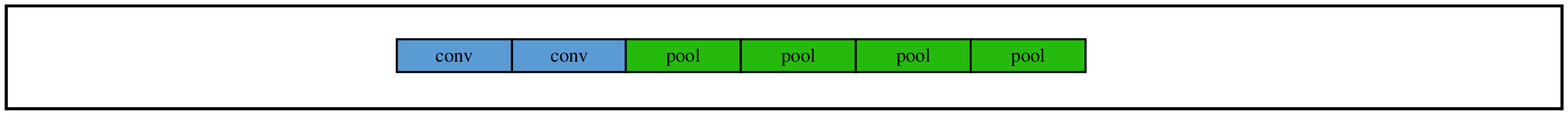}%
		\label{fig_velocity_4}}
	\caption{An example to illustrate the proposed $x$-reference velocity updating method.}
	\label{fig_velocity}
\end{figure*}

For an intuitive understanding the proposed $x$-reference velocity updating method, an example is provided in Fig.~\ref{fig_velocity}. Specifically, Fig.~\ref{fig_velocity_1} displays the $gBest$ and $x$ that are used to do the ``global search'' part in the velocity updating. In Fig.~\ref{fig_velocity_2}, the convolutional layers and pooling layers are collected from $gBest$ and $x$. Because the lengths of convolutional layers and pooling layers from $x$ are $2$ and $4$, and those from $gBest$ are $3$ and $2$, the last convolutional layer from the convolutional layer part from $gBest$ is truncated, and the other two pooling layers are padded to the tail of the pooling layer part of $gBest$. In particular, the padded pooling layers are created with the values of encoded information equal to $0$. Fig.~\ref{fig_velocity_3} demonstrates the updating between the convolutional layer part and pooling layer part from $gBest$ and $x$. Fig.~\ref{fig_velocity_4} shows the results of this updating.

No matter whether padding or truncating is operated in the designed $x$-reference velocity updating method, the goal is to make the same length of convolutional layers and pooling layers in $gBest$ to that in $x$, respectively. The mechanism behind this design is discussed as follows. In PSOAO, there are a group of particles with different lengths in the population, with the same goal of searching for the optimal architectures of FCAEs for solving image classification tasks. If we have each particle follow the length of $gBest$ (i.e., the $gBest$-reference velocity updating method), all the particles will have the same length to that of $gBest$ from the second generation. Because the $pBest_i$ is chosen from the memory of each particle, the $gBest$, $pBest_i$, and current particle $x$ may all have the same length from the third generation. Consequently, all particles participate in the optimization with one particular depth of FCAE and only change the encoded information. Indeed, the variation of length regarding $gBest$ can be seen as an exploration search behavior, while that of the encoded information is viewed as an exploitation search behavior. When all particles are in the same length, the length of $gBest$ will be constant until it terminates. In this regard, the exploration search ability is lost if we employ the $gBest$-reference velocity updating method. In addition, keeping the length of $x$ equal to $gBest$ can also be viewed as the losses of diversity, which would easily lead to the premature convergence in population-based algorithms. Both the loss of exploration search and the premature phenomenon will result in a poor performance. An experiment is conducted in Subsection~\ref{sec_5_4} to further quantitatively investigate this velocity updating design.

\subsection{Deep Training on $gBest$}
When the evolution of PSOAO is finished, the best particle, $gBest$, is picked for deep training. As stated in Subsection~\ref{sec_fitness} that each particle is trained with only a few epochs, which is not the optimal performance for solving real-world applications.  To solve this concern, the deep training is necessary. Typically, the process of the deep training is the same to the fitness evaluation in Subsection~\ref{sec_fitness} except for a larger epoch number, say 100 or 200.

\section{Experimental Design}
\label{section_4}
In this section, the benchmark datasets, peer competitors, and the parameter settings are detailed for the experiments investigating the performance of the proposed FCAE in which its architecture is optimized by the proposed PSOAO algorithm.
\subsection{Benchmark Datasets}
\label{sec_4_1}
\begin{figure}
	\centering
	\subfloat[Examples from the CIFAR-10 dataset. From left to right, they are from the categories of truck, ship,  horse, frog,  dog,  deer,  cat,  bird, automobile and airplane, respectively.]{\includegraphics[width=0.9\columnwidth]{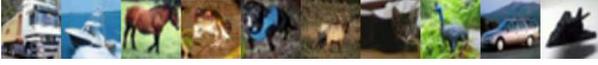}%
	\label{fig_dataset_cifar}}
	\hfill
	\subfloat[Examples from the Mnist dataset. From left to right, they are from the categories of $0\cdots9$, respectively.]{\includegraphics[width=0.9\columnwidth]{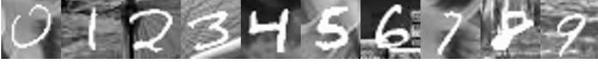}%
	\label{fig_dataset_mnist}}
	\hfill
	\subfloat[Examples from the STL-10 dataset. From left to right, they are from categories of airplane, bird, car, cat, deer, dog, horse, monkey, ship and truck, respectively.]{\includegraphics[width=0.9\columnwidth]{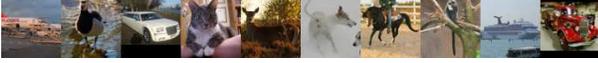}%
	\label{fig_dataset_stl}}
	\hfill
	\subfloat[Examples from the Caltech-101 dataset. From left to right, they are from categories of accordio, bass, camera, dolphin, elephant, ferry, gramophone, headphone, lamp and sunflower, respectively.]{\includegraphics[width=0.9\columnwidth]{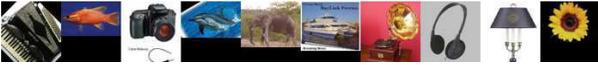}%
	\label{fig_dataset_caltech101}}
	\caption{Examples from the chosen benchmark datasets.}
	\label{fig_dataset}
\end{figure}

The experiments are conducted on four image classification benchmark datasets, which are widely used and specifically for investigating the performance of AEs. They are the CIFAR-10 dataset~\cite{krizhevsky2009learning}, the MNIST dataset~\cite{lecun1998gradient}, the STL-10 dataset~\cite{coates2011analysis}, and the Caltech-101 dataset~\cite{fei2007learning}. Fig.~\ref{fig_dataset} shows examples from these benchmark datasets. In the following, these chosen datasets are briefly introduced.

\subsubsection{CIFAR-10 dataset} It contains $50,000$ training images and $10,000$ test images Each one is a 3-channel RGB image in the size of $32\times 32$, and belongs to one of $10$ categories of natural objectives (i.e., truck, ship, horse, frog, dog, deer, cat, bird, automobile, and airplane). Each category is with roughly the same number of images. In addition, different objects occupy different areas of the images.

\subsubsection{MNIST dataset} It is a handwritten digit recognition dataset to classify the numeral numbers of $0\cdots9$, including $60,000$ training images and $10,000$ test images. Each one is a 1-channel gray image in the size of $28\times 28$, and the samples in each category are with different variations, such as rotations. Each category is composed of the same number of image samples.

\subsubsection{STL-10 dataset} It is a widely used dataset for unsupervised learning, containing $100,000$ unlabeled images, $5,000$ training images, and $8,000$ test images from $10$-category natural image object recognition (i.e., airplane, bird, car, cat, deer, dog, horse, monkey, ship, truck). Each one is a 3-channel RGB image in the size of $96\times 96$. In addition, the unlabeled images contain images that are beyond the $10$ categories. Due to the small number of training samples, this dataset challenges the feature learning ability of CAEs/AEs and PSOAO.

\subsubsection{Caltech-101 dataset} It is a $101$-category image classification dataset where the weights and heights of images vary from $80$ to $708$ pixels. Most images are 3-channel RGB while occasionally gray, and with different images in each category from $31$ to $800$. In addition, most images only display a small part of the image, and other areas are occupied by noises for increasing the difficulty in classification. Due to the quite small number of images, and the non-identical numbers of images in each category, this dataset also challenges the feature learning algorithms.

\subsection{Peer Competitors}
\label{sec_4_2}
Peer competitors for the proposed FCAE, which have been introduced in Section~\ref{section_1}, are employed for performing the comparisons on the chosen image classification benchmark datasets. They are the CAE~\cite{masci2011stacked}, Convolutional RBM (CRBM)~\cite{lee2009convolutional}, and the state-of-the-art Convolutional Denoising AE (CDAE)~\cite{du2017stacked}. In addition, two widely used variants of AEs are also employed as the peer competitors for a comprehensive comparison. They are the Sparse AE (SAE)~\cite{shin2013stacked} and Denoising AE (DAE)~\cite{vincent2010stacked}.

Because this paper aims at proposing an FCAE that could be stacked to the state-of-the-art CNNs, peer competitors for comparisons here also include the stacked forms of these CAEs/AEs, i.e., the Stacked CAE (SCAE), the Stacked CRBM (SCRBM), the Stacked CDAE (SCDAE), the Stacked SAE (SSAE), and the stacked DAE (SDAE). The Stacked form of the proposed FCAE is SFCAE.
\subsection{Parameter Settings}
\label{sec_4_3}
The peer competitors SCAE, SCRBM, SSAE, and SDAE have been investigated very recently on the chosen benchmark datasets, and their architectures have been manually tuned with domain expertise~\cite{du2017stacked}. Their classification results are directly cited from the original publications, thus none of their parameter settings needs to be specified. In addition, the state-of-the-art SCDAE provides the classification results with the usage of only one and two building blocks on the chosen benchmark datasets. In order to do a fair comparison, we also perform experiments on SFCAE with at most two building blocks. Note that the SFCAE is tested on the chosen benchmark datasets without the preprocessing of data augmentation for keeping consistency to its peer competitors. In the following, the parameter settings of PSOAO are provided in detail.

In the designed PSOAO algorithm, PSO related parameters are specified based on their conventions~\cite{bratton2007defining}, i.e., the inertia weight $w$ is set to be $0.72984$, the acceleration constants $c_1$ and $c_2$ are both set to be $1.496172$, and the initial velocity is set to be $0$. The training sets of MNIST and CIFAR-10, the unlabeled data of STL-10 are naturally used for their fitness evaluations. Due to the non-identical and a quite small number of images in each category of the Caltech-101 dataset, $30$ images randomly selected from each category are used for the fitness evaluations and also as the training set based on the suggestions in~\cite{du2017stacked}. Because inappropriate settings of convolutional operations and pooling operations would lead to an unaffordable computational cost and make FCAE being incompetitive, in the exploration of each particle, the number of feature maps is set to be $[20, 100]$, the kernel with the same size of width and height is set to be $[2,5]$, the maximal number of pooling layers is set to be $1$, and that of convolutional layer is set to be $5$. Note that only the square convolutional filters and pooling kernels are investigated in the experiments, which is based on the conventions from state-of-the-art CNNs~\cite{he2016deep,huang2017densely}. In addition, the coefficient of $l_2$ term is set to be $[0.0001,0.01]$, which is a commonly utilized range for training neural networks in practice.

FCAE with the architecture determined by PSOAO and the weights initialized by the widely used  Xavier method~\cite{glorot2010understanding}, by adding one full connection layer with $512$ units and 50\% Dropout~\cite{srivastava2014dropout} from the conventions of deep learning community, are used for the deep training. We investigate the classification results by feeding the trained model with the corresponding test set\footnote{The test data of Caltech-101 dataset is the entire dataset excluding from the training set.}, employing the widely used rectifier linear unit~\cite{glorot2011deep} as the activation function, the Adam~\cite{kingma2014adam} optimizer with its default settings as the training algorithm for weight optimization, and the BatchNorm~\cite{ioffe2015batch} technique for speeding up the training. For keeping consistency in the results to be compared, the experiments with the trained model on each benchmark dataset are also independently performed $5$ times. Due to the extreme imbalance training data exist in the Caltech-101 dataset, we investigate this dataset based on its convention~\cite{lee2009convolutional}, i.e., investigating the classification accuracy on each category of the images, and then reporting the mean and standard derivations over the whole dataset.

The proposed PSOAO algorithm is implemented by Tensorflow~\cite{abadi2016tensorflow}, and each copy of the source code runs on one GPU card with the same model number of GTX1080. In addition, the architecture configurations of FCAEs optimized by PSOAO for the benchmark datasets in these experiments are provided in Subsection~\ref{sub_section_architecture}. Training time of the proposed PSOAO algorithm on the benchmark datasets are shown in Table~\ref{table_comsuming_time_alg}.
\begin{table}[htp]
	\renewcommand{\arraystretch}{1.7}
	\caption{Consumed time (hours) of the proposed PSOAO algorithm for different benchmark datasets.}
	\label{table_comsuming_time_alg}
	\begin{center}
		\begin{tabular}{cccc}
			\hline
			
			\hline
			\textbf{CIFAR-10} &  \textbf{MNIST} & \textbf{STL-10} & \textbf{Caltech-101} \\
			\hline
			
			\hline
			 81.5 & 118&230 & 22.5\\
			\hline
			
			\hline
		\end{tabular}
	\end{center}
\end{table}

\section{Experimental Results and Analysis}
\label{section_5}
\subsection{Overview Performance}
\label{sec_5_1}
Table~\ref{table_overview_performance} shows the mean and standard derivations of the classification accuracies of the proposed FCAE method, whose architecture is optimized by the designed PSOAO algorithm, and the peer competitors on the benchmark datasets. Because literatures do not provide the standard derivations of SCAE on CIFAR-10 and MNIST, SCRBM on CIFAR-10, and SDAE on MNIST, only their mean classification accuracies are shown. The references in Table~\ref{table_overview_performance} denote the sources of the corresponding mean classification accuracies, and the best mean classification results are highlighted in bold. The terms ``SFCAE-1'' and ``SFCAE-2'' refer to the SFCAE with one and two building blocks, respectively, which is the same meaning as the terms ``SCDAE-1'' and ``SCDAE-2.''

\begin{table}[htp]
	\renewcommand{\arraystretch}{1.7}
	\caption{The classification accuracy of the proposed FCAE method against its peer competitors on the chosen benchmark datasets.}
	\label{table_overview_performance}
	
	\begin{center}
		\begin{tabular}{ccccc}
			\hline
			
			\hline
			\textbf{Algorithm} & \textbf{CIFAR-10} &  \textbf{MNIST} & \textbf{STL-10} & \textbf{Caltech-101} \\
			\hline
			
			\hline
			SSAE & $74.0$ ($0.9$) & $96.29$ ($0.12$) & $55.5$ ($1.2$) & $66.2$ ($1.2$) \\
			\hline
			SDAE & $70.1$ ($1.0$) & $99.06$~\cite{vincent2010stacked} & $53.5$ ($1.5$) & $59.5$ ($0.3$) \\
			\hline
			SCAE & $78.2$~\cite{masci2011stacked} & $99.29$~\cite{masci2011stacked} & $40.0$ ($3.1$) & $58.0$ ($2.0$)\\
			\hline
			SCRBM & $78.9$~\cite{krizhevsky2010convolutional} & $99.18$~\cite{lee2009convolutional} & $43.5$ ($2.3$) & $65.4$ ($0.5$)~\cite{lee2009convolutional} \\
			\hline
			SCDAE-1 & $75.0$ ($1.2$) & $99.17$ ($0.10$) & $56.6$ ($0.8$) & $71.5$ ($1.6$) \\
			\hline
			SCDAE-2 & $80.4$ ($1.1$) & $99.38$ ($0.05$) &$60.5$ ($0.9$) & $78.6$ ($1.2$) \\
			\hline
			SFCAE-1 & $78.9$ ($0.3$) & $99.30$ ($0.03$) & $\textbf{61.2 (1.2)}$ & $\textbf{79.8 (0.0)}$ \\
			\hline
			SFCAE-2 & $\textbf{83.5 (0.5)}$ & $\textbf{99.51 (0.09)}$ & $56.8$ ($0.2$) & $79.6$ ($0.0$) \\
			\hline
			
			\hline
		\end{tabular}
	\end{center}
\end{table}

It is clearly shown in Table~\ref{table_overview_performance} that FCAE outperforms the traditional AEs (i.e., the SSAE and the SDAE) and traditional CAEs (i.e., the SCAE and the SCRBM) on all benchmark datasets. In addition, FCAE also outperforms the state-of-the-art CAEs (i.e, the SCDAE-1 and the SCDAE-2) on these benchmark datasets. Furthermore, the best results on CIFAR-10 and MNIST are reached by SFCAE-2, and those on STL-10 and Caltech-101 are by SFCAE-1. Note that SFCAE-2 performs worse on STL-10 and Caltech-101 than SFCAE-1, which is caused by the much smaller numbers of training instances in these two benchmark datasets, and deeper architectures are suffered from the overfitting problem. Because CIFAR-10 and MNIST are with many more training samples ($50,000$ in CIFAR-10 and $60,000$ in MNIST), a deeper architecture naturally results in the promising classification accuracy. In summary, when the architecture of the proposed FCAE method is optimized by the designed PSOAO algorithm, FCAE shows superior performance among its peer competitors on all the four image classification benchmark datasets.

\subsection{Evolution Trajectory of PSOAO}
\label{sec_5_2}
\begin{figure}
	\centering
	\subfloat[The evolution trajectories on the CIFAR-10 dataset. ]{\includegraphics[width=\columnwidth]{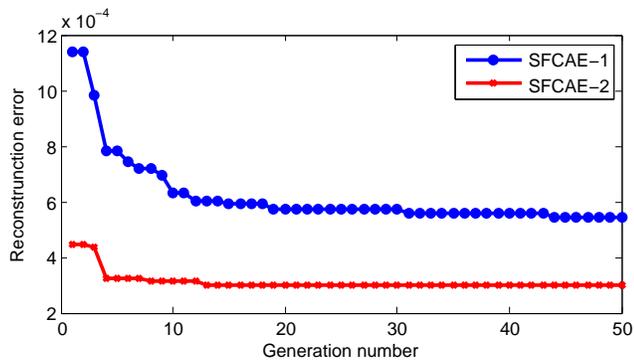}%
		\label{fig_cifar_his}}
	\hfill
	\subfloat[The evolution trajectories on the MNIST dataset. ]{\includegraphics[width=\columnwidth]{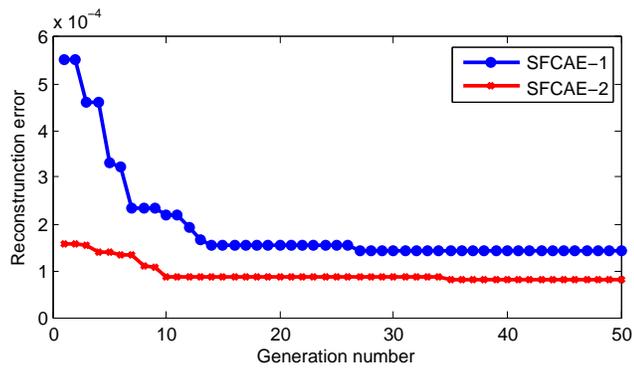}%
		\label{fig_mnist_his}}
	\hfill
	\subfloat[The evolution trajectories on the STL-10 dataset. ]{\includegraphics[width=\columnwidth]{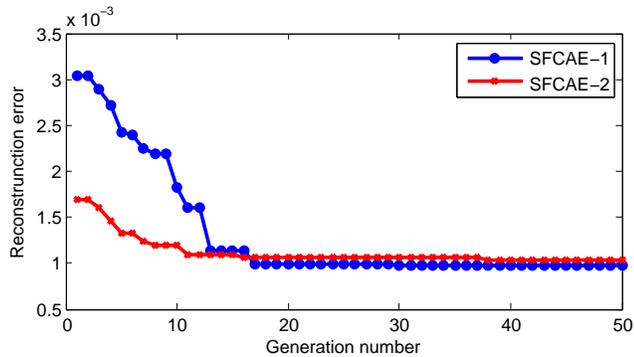}%
		\label{fig_stl_his}}
	\hfill
	\subfloat[The evolution trajectories on the Caltech-101 dataset. ]{\includegraphics[width=\columnwidth]{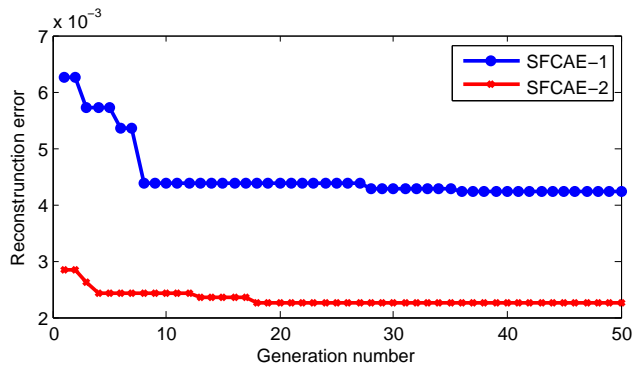}%
		\label{fig_caltech_his}}
	\caption{Trajectories of the PSOAO algorithm in automatically discovering the architectures of FCAE on the chosen benchmark datasets.}
	\label{fig_his}
\end{figure}
In order to intuitively investigate the efficacy of the designed PSOAO algorithm in optimizing the architectures of the proposed FCAE method, its evolution trajectories on the chosen benchmark datasets during the training phases are plotted in Fig.~\ref{fig_his}, where the horizontal axis denotes the number of generations in the evolution, and the vertical axis denotes the fitness values of the $gBest$.

As can be seen from Figs.~\ref{fig_cifar_his},~\ref{fig_mnist_his},~\ref{fig_stl_his}, and~\ref{fig_caltech_his}, PSOAO has converged within the specified maximal generation number. Specifically, it has converged since about the $15$th generation on all benchmark datasets for both SFCAE-1 and SFCAE-2, and about the $5$th generation on the CIFAR-10 and Caltech-101 datasets for SFCAE-2. Note that the reconstruction error of SFCAE-2 is smaller than that of SFCAE-1 on STL-10 dataset (shown in Fig.~\ref{fig_stl_his}), which is caused by the different input data for them.

\subsection{Performance on Different Numbers of Training Examples}
\label{sec_5_3}
In this subsection, we investigate the classification performance of the proposed FCAE method whose architecture is optimized by the designed PSOAO algorithm on different numbers of training samples. The peer competitors on the MNIST dataset are SCRBM~\cite{lee2009convolutional}, ULIFH~\cite{huang2007unsupervised},  SSE~\cite{weston2012deep}, and SCAE~\cite{masci2011stacked}, and those for the CIFAR-10 dataset are SCAE~\cite{masci2011stacked}, Mean-cov. RBM~\cite{krizhevsky2010convolutional}, and  K-means (4k feat)~\cite{coates2011analysis}. The reason for choosing these benchmark datasets and peer competitors is that the literature has provided their corresponding information that is widely used by the comparisons between various variants of CAEs.

\begin{table}[htp]
	\renewcommand{\arraystretch}{1.7}
	\caption{The classification accuracy of FCAE-2 against peer competitors on the different numbers of training samples from MNIST.}
	\label{table_number_mnist}
	
	\begin{center}
		\begin{tabular}{cccccccc}
			\hline
			
			\hline
			\textbf{\# samples} & 1K &  2K & 3K & 5K & 10K  & 60K \\
			\hline
			
			\hline
			SCRBM &$97.38$ & $97.87$ & $98.09$ & $98.41$ &---  & $99.18$ \\
			\hline
			ULIFH & $96.79$ & $97.47$ & --- & $98.49$ & ---  & $99.36$ \\
			\hline
			SSAE &$97.27$ & --- & $98.17$ & --- & ---  & $98.50$ \\
			\hline
			SCAE & $92.77$ &---&---&---&$98.12$  & $99.29$ \\
			\hline
			SFCAE& \textbf{97.49} & \textbf{98.45} & \textbf{98.80} & \textbf{99.09} & \textbf{99.16} & \textbf{99.51} \\
			\hline
			
			\hline
		\end{tabular}
	\end{center}
\end{table}

\begin{table}[htp]
	\renewcommand{\arraystretch}{1.7}
	\caption{The classification accuracy of FCAE-2 against peer competitors on the different numbers of training samples from CIFAR-10.}
	\label{table_number_cifar}
	
	\begin{center}
		\begin{tabular}{cccc}
			\hline
			
			\hline
			\textbf{\# samples} & 1K & 10K  & 50K \\
			\hline
			
			\hline
			SCAE & $47.70$ & $65.65$ &$78.20$ \\
			\hline
			Mean-cov. RBM & --- & --- & $71.00$ \\
			\hline
			K-means (4k feat) & --- & --- & $79.60$ \\
			\hline
			SFCAE& \textbf{53.79} & \textbf{73.96} & \textbf{83.47} \\
			\hline
			
			\hline
		\end{tabular}
	\end{center}
\end{table}

Tables~\ref{table_number_mnist} and \ref{table_number_cifar} show the experimental results of FCAE-2 with the architecture confirmed in Subsection~\ref{sec_5_1} on different numbers of training samples of the MNIST and CIFAR-10 benchmark datasets. The symbol  ``---'' denotes that there is no result reported in the corresponding literature. The best classification accuracy is highlighted in bold.

As can be seen from Tables~\ref{table_number_mnist} and \ref{table_number_cifar}, SFCAE-2 surpasses all peer competitors on these two datasets. Especially, with a much smaller number (1K) of training examples, SFCAE-2 performs better than the seminal work of CAE (SCAE) with $4.72\%$ classification accuracy improvement on MNIST and $6.09\%$ on CIFAR-10. These results show the promising scalability of SFCAE-2 in dealing with different numbers of training samples.

\subsection{Investigation on $x$-reference Velocity Calculation}
\label{sec_5_4}

\begin{figure}[!htp]
	\centering
	\subfloat[Classification accuracy of SFCAE-1 by truncating $gBest$ and $x$.]{\includegraphics[width=\columnwidth]{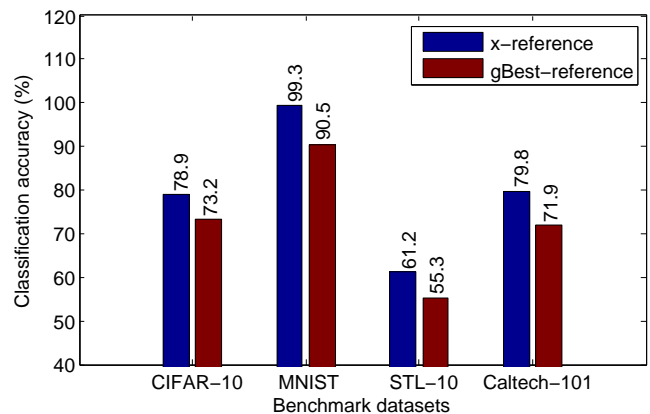}%
		\label{fig_truncated_1}}
	\hfill
	\subfloat[Classification accuracy of SFCAE-2 by truncating $gBest$ and $x$.]{\includegraphics[width=\columnwidth]{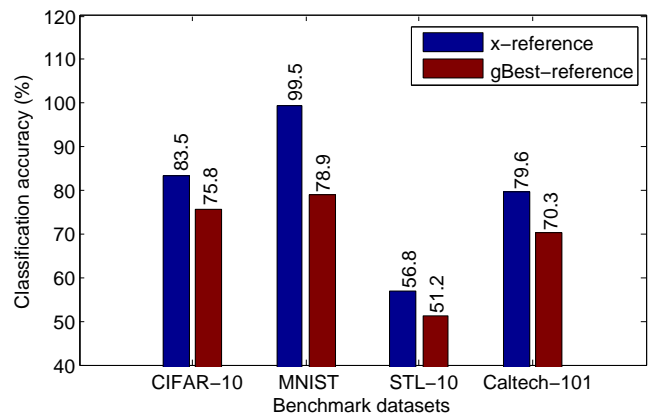}%
		\label{fig_truncated_2}}
	\caption{Classification accuracy comparisons between the $x$-reference and $gBest$-reference velocity updating strategy in the designed PSOAO method.}
	\label{fig_truncated_c}
\end{figure}

To further investigate the superiority of the proposed $x$-reference velocity updating method, we replace it from the proposed PSOAO algorithm with the $gBest$-reference velocity updating method, to compare the performance on the chosen benchmark datasets introduced in Subsection~\ref{sec_4_1}. To achieve this, we first let the length of $pBest_i$ equal to that of $gBest$. If the length of the convolutional layers in $pBest_i$ is less than that of $gBest$, zeros are padded. Otherwise, truncating the corresponding parts from $pBest_i$. We use this method in the pooling layers of $pBest_i$ and these two types of layers in $x$. Then, we use Equations~(\ref{equ_veclocity_calculate}) and (\ref{equ_position_update}) to update the position of each particle. The experimental results are shown in Fig.~\ref{fig_truncated_c}, where Fig.~\ref{fig_truncated_1} displays the results of FSCAE-1 while Fig.~\ref{fig_truncated_2} displays those of FSCAE-2.

As can be seen from Fig.~\ref{fig_truncated_1}, with the $x$-reference velocity updating method, SFCAE-1 achieves the classification accuracy improvements of $5.7\%$, $8.8\%$, $5.9\%$, and $7.9\%$ on the CIFAR-10, MNIST, STL-10, and Caltech-101 benchmark datasets, respectively. The same promising performance of SFCAE-2 can also be observed with the classification accuracy improvements of $7.7\%$, $10.6\%$, $5.6\%$, and $9.6\%$ on these chosen benchmark datasets, as shown in Fig.~\ref{fig_truncated_2}.

In summary, the analysis and experimental results convincely justify the effectiveness of the proposed $x$-reference velocity updating method in the proposed PSOAO algorithm.

\subsection{Investigation on The Obtained architectures}
\label{sub_section_architecture}

Tables~\ref{table_architecture_cifar10_1},~\ref{table_architecture_mnist_1},~\ref{table_architecture_stl10_1}, and~\ref{table_architecture_caltech_1} show the architecture configurations of SFCAE-1 for the CIFAR-10, MNIST, STL-10 and Caltech-101 datasets, respectively. Additionally, Tables~\ref{table_architecture_cifar10_2},~\ref{table_architecture_mnist_2},~\ref{table_architecture_stl10_2}, and~\ref{table_architecture_caltech_2} show the architecture configurations of SFCAE-2 for the CIFAR-10, MNIST, STL-10 and Caltech-101 datasets, respectively. Noting that the SAME convolutional layers, the max pooling layers and the strides with $1\times 1$ are used based on the experimental settings, therefore, these configurations are not shown in these tables.

\begin{table}[!htp]
	\caption{The architecture configuration of SFCAE-1 on the CIFAR-10 dataset.}
	\label{table_architecture_cifar10_1}
	\renewcommand{\arraystretch}{1.0}
	
	\begin{center}
		\begin{tabular}{cc}
			\hline
			
			\hline
			\textbf{layer type} & \textbf{configuration} \\
			\hline
			
			\hline
			conv & Filter: $2\times 2$,  \# Feature map: $24$, L$_2$: $0.0016$ \\
			conv & Filter: $3\times 3$,  \# Feature map: $57$, L$_2$: $0.0001$ \\
			conv & Filter: $5\times 5$,  \# Feature map: $63$, L$_2$: $0.0096$\\
			conv & Filter: $5\times 5$,  \# Feature map: $35$, L$_2$: $0.0071$\\
			conv & Filter: $3\times 3$,  \# Feature map: $76$, L$_2$: $0.0015$\\
			pooling & Kernel: $2\times2$, Stride: $2\times 2$\\
			\hline
			
			\hline		
		\end{tabular}
	\end{center}

\end{table}

\begin{table}[!htp]
	\caption{The architecture configuration of SFCAE-2 on the CIFAR-10 dataset.}
	\label{table_architecture_cifar10_2}
	\renewcommand{\arraystretch}{1.0}
	
	\begin{center}
		\begin{tabular}{cc}
			\hline
			
			\hline
			\textbf{layer type} & \textbf{configuration} \\
			\hline
			
			\hline
			conv & Filter: $4\times 4$,  \# Feature map: $36$, L$_2$: $0.0001$ \\
			pooling & Kernel: $2\times2$, Stride: $2\times 2$\\
			\hline
			
			\hline		
		\end{tabular}
	\end{center}

\end{table}

\begin{table}[!htp]
	\caption{The architecture configuration of SFCAE-1 on the MNIST dataset.}
	\label{table_architecture_mnist_1}
	\renewcommand{\arraystretch}{1.0}
	
	\begin{center}
		\begin{tabular}{cc}
			\hline
			
			\hline
			\textbf{layer type} & \textbf{configuration} \\
			\hline
			
			\hline
			conv & Filter: $2\times 2$,  \# Feature map: $100$, L$_2$: $0.0010$ \\
			conv & Filter: $2\times 2$,  \# Feature map: $82$, L$_2$: $0.0018$ \\
			conv & Filter: $2\times 2$,  \# Feature map: $100$, L$_2$: $0.0001$\\
			conv & Filter: $2\times 2$,  \# Feature map: $100$, L$_2$: $0.0001$\\
			pooling & Kernel: $2\times2$, Stride: $2\times 2$\\
			\hline
			
			\hline		
		\end{tabular}
	\end{center}

\end{table}

\begin{table}[!htp]
	\caption{The architecture configuration of SFCAE-2 on the MNIST dataset.}
	\label{table_architecture_mnist_2}
	\renewcommand{\arraystretch}{1.2}
	
	\begin{center}
		\begin{tabular}{cc}
			\hline
			
			\hline
			\textbf{layer type} & \textbf{configuration} \\
			\hline
			
			\hline
			conv & Filter: $3\times 3$,  \# Feature map: $89$, L$_2$: $0.0001$ \\
			conv & Filter: $3\times 3$,  \# Feature map: $90$, L$_2$: $0.0001$ \\
			conv & Filter: $3\times 3$,  \# Feature map: $93$, L$_2$: $0.0005$\\
			pooling & Kernel: $2\times2$, Stride: $2\times 2$\\
			\hline
			
			\hline		
		\end{tabular}
	\end{center}

\end{table}

\begin{table}[!htp]
	\caption{The architecture configuration of SFCAE-1 on the STL-10 dataset.}
	\label{table_architecture_stl10_1}
	\renewcommand{\arraystretch}{1.0}
	
	\begin{center}
		\begin{tabular}{cc}
			\hline
			
			\hline
			\textbf{layer type} & \textbf{configuration} \\
			\hline
			
			\hline
			conv & Filter: $2\times 2$,  \# Feature map: $85$, L$_2$: $0.0096$ \\
			conv & Filter: $3\times 3$,  \# Feature map: $77$, L$_2$: $0.0086$ \\
			conv & Filter: $3\times 3$,  \# Feature map: $83$, L$_2$: $0.0025$\\
			pooling & Kernel: $2\times2$, Stride: $2\times 2$\\
			\hline
			
			\hline		
		\end{tabular}
	\end{center}

\end{table}

\begin{table}[!htp]
	\caption{The architecture configuration of SFCAE-2 on the STL-10 dataset.}
	\label{table_architecture_stl10_2}
	\renewcommand{\arraystretch}{1.0}
	
	\begin{center}
		\begin{tabular}{cc}
			\hline
			
			\hline
			\textbf{layer type} & \textbf{configuration} \\
			\hline
			
			\hline
			conv & Filter: $2\times 2$,  \# Feature map: $83$, L$_2$: $0.0094$ \\
			conv & Filter: $4\times 4$,  \# Feature map: $49$, L$_2$: $0.0087$ \\
			pooling & Kernel: $2\times2$, Stride: $2\times 2$\\
			\hline
			
			\hline		
		\end{tabular}
	\end{center}

\end{table}

\begin{table}[!htp]
	\caption{The architecture configuration of SFCAE-1 on the Caltech-101 dataset.}
	\label{table_architecture_caltech_1}
	\renewcommand{\arraystretch}{1.0}
	
	\begin{center}
		\begin{tabular}{cc}
			\hline
			
			\hline
			\textbf{layer type} & \textbf{configuration} \\
			\hline
			
			\hline
			conv & Filter: $2\times 2$,  \# Feature map: $14$, L$_2$: $0.0084$ \\
			conv & Filter: $5\times 5$,  \# Feature map: $8$, L$_2$: $0.0002$ \\
			pooling & Kernel: $2\times2$, Stride: $2\times 2$\\
			\hline
			
			\hline		
		\end{tabular}
	\end{center}

\end{table}

\begin{table}[!htp]
	\caption{The architecture configuration of SFCAE-2 on the Caltech-101 dataset.}
	\label{table_architecture_caltech_2}
	\renewcommand{\arraystretch}{1.0}
	
	\begin{center}
		\begin{tabular}{cc}
			\hline
			
			\hline
			\textbf{layer type} & \textbf{configuration} \\
			\hline
			
			\hline
			conv & Filter: $5\times 5$,  \# Feature map: $15$, L$_2$: $0.0051$ \\
			pooling & Kernel: $2\times2$, Stride: $2\times 2$\\
			\hline	
			
			\hline
		\end{tabular}
	\end{center}

\end{table}

The obtained architectures are based on our designed encoding strategy. The encoding strategy in this paper is designed based on the architectures of the traditional convolutional neural networks (CNNs) that are composed of several blocks, and each block is composed of several convolutional layers followed by the pooling layer. Unsurprisingly, these obtained architectures follow the architectures of the traditional CNNs and the architectures of known and manually designed network, such as the VGGNet~\cite{simonyan2014very}. 

Recently, there are two famous types of CNNs, i.e., ResNet~\cite{he2016deep} and DensNet~\cite{huang2017densely}. Because these two networks are different from the architectures of the traditional CNNs due to the skip and dense connections, the obtained architectures are different to these networks. Due to the promising performance of ResNet and DenseNet on large-scale image classification tasks shown in their experiments, we will in future investigate a new encoding strategy that is capable of encoding the skip connections of ResNet and dense connections of DensNet.

\section{Conclusions and Future Work}
\label{section_7}

The goal of the paper is to develop a novel PSO algorithm (namely PSOAO) to automatically discover the optimal architecture of the flexible convolutional auto-encoder (named FCAE) for image classification problems without manual intervention. This goal has been successfully achieved by defining the FCAE that has the potential to construct the state-of-the-art deep convolutional neural networks, designing an efficient encoding strategy that is capable of representing particles with non-identical lengths in PSOAO, and developing an effective velocity updating mechanism for these particles. The FCAE with the ``optimal'' architecture is achieved by PSOAO, and compared with five peer competitors including the most state-of-the-art algorithms on four benchmark datasets, specifically used by auto-encoders for image classification. The experimental results show that FCAE remarkably outperforms all the compared algorithms on all adopted benchmark datasets in term of their classification accuracies. Furthermore, FCAE with only one building block can surpass the state-of-the-art with two building blocks on the STL-10 and Caltech-101 benchmark datasets. In addition, FCAE reaches the best classification accuracies compared with the four peer competitors when only 1K, 2K, 3K, 5K, and 10K training images of the MNIST benchmark dataset are used, and significantly outperforms three peer competitors when only 1K and 10K training images of the CIFAR-10 benchmark dataset are used. Moreover, the proposed PSOAO algorithm also shows the excellent characteristic of fast convergence by investigating its evolution trajectories, and the effective velocity updating mechanism through the quantitative comparison to its opponents. 

Although deep CNNs have achieved the current state-of-the-art results for image classification, the architectures and hyper-parameters are largely determined by manual tuning based on domain expertise/knowledge. This paper provides a direction, which shows that such manual work can be replaced by automatic learning through evolutionary approaches. In the future, we will investigate simpler evolutionary methods on more complex CNN models with using much less computational resources.

\ifCLASSOPTIONcaptionsoff
  \newpage
\fi

\end{document}